# Neural network modelling of kinematic and dynamic features for signature verification

Moises Diaz, Miguel A. Ferrer, Jose Juan Quintana, Adam Wolniakowski, Roman Trochimczuk, Konstantsin Miatliuk, Giovanna Castellano, Gennaro Vessio

## ABSTRACT

Online signature parameters, which are based on human characteristics, broaden the applicability of an automatic signature verifier. Although kinematic and dynamic features have previously been suggested, accurately measuring features such as arm and forearm torques remains challenging. We present two approaches for estimating angular velocities, angular positions, and force torques. The first approach involves using a physical UR5e robotic arm to reproduce a signature while capturing those parameters over time. The second method, a cost-effective approach, uses a neural network to estimate the same parameters. Our findings demonstrate that a simple neural network model can extract effective parameters for signature verification. Training the neural network with the MCYT300 dataset and cross-validating with other databases, namely, BiosecurID, Visual, Blind, OnOffSigDevanagari-75 and OnOffSigBengali-75 confirm the model's generalization capability. The trained model is available at: https://github.com/gvessio/SignatureKinematics.

## 1. Introduction

An *online signature* is typically represented by the parametric equations of its trajectory, denoted as $(x(t), y(t))$, which are captured when the signing tool contacts the digital surface. Additionally, some digitizers capture other function-based parameters, such as the vertical pressure exerted by the pen tip, azimuthal and altitude angles of the pen, and even the pen's in- air trajectory. As a physiological biometric trait, a signature is used in various applications, including access control, com- mercial transactions, document forgery detection, and the provision of evidence in legal scenarios such as the verification of last wills [9]. In biometrics, where impostors may attempt to forge signatures with varying degrees of skill, robust verification methods are crucial. Since the execution of a signature inherently involves movements of the hand, arm, and forearm, it is hypothesized that these motions may contain kinematic and dynamic unique characteristic of the signer [7]. From a kinematic perspective, this action can be characterized by the arm's angular position, $\theta(t)$, and angular velocity, $\omega(t)$. Dynamically, these movements are facilitated by force torques, $\tau(t)$, applied at the joints.

One method used to obtain this valuable biomechanical in- formation involves a physical robot programmed to mimic the act of signing. While a robot's ability to accurately replicate these movements depends on its configuration, working area, and degrees of freedom, it can effectively capture kinematic and dynamic features during the process. However, accessing these robots is costly and cumbersome. To improve access to robotic features, our contributions are:

- The angular positions, $\theta_r(t) \in (\theta_1(t), ..., \theta_6(t))$, angular velocities, $\omega_r(t) \in (\omega_1(t), ..., \omega_6(t))$, and force torques, $\tau_r(t) \in (\tau_1(t), ..., \tau_6(t))$ of a UR5e robotic were recorded during the execution of the 16,500 online signatures from the MCYT330 dataset (DS1).

- We trained a multilayer perceptron (MLP)-based neural network to estimate the kinematic and dynamic features $\hat{\theta}(t)$, $\hat{\omega}(t)$, $\hat{\tau}(t)$ by taking the trajectory of a signature $(x(t), y(t))$ as input.

- We conducted a two-fold experiment: first, we assessed the MLP's estimation capacity, and secondly, we compared the performance of features generated by the UR5e robot against those produced by the MLP in a dynamic time warping (DTW)-based automatic signature verifier (ASV). Additionally, we demonstrated the neural network's generalization capability by estimaing effective kinematic and dynamic features in five third-party databases, namely, BiosecurID (DS2), Visual (DS3), Blind (DS4), OnOffSigDevanagari-75 (DS5) and OnOffSigBengali-75 (DS6).

Figure 1 illustrates the methodology employed in this study. The trained MLP model is publicly available for research and practical applications.

The rest of this paper is organized as follows. Section 2 reviews prior work. Section 3 details the proposed method. Section 4 presents the signature verification experiments con- ducted. Section 5 concludes the article.

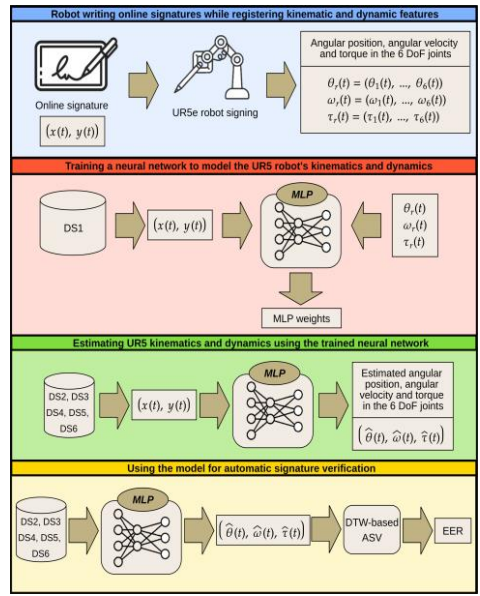

Fig. 1: Overview of the proposed method.

## 2. Related work

### 2.1. Robots replicating human writing trajectories

Various robotic arms have been developed to replicate human writing trajectories, drawing significant interest from the art world and at technical exhibitions. For example, the Paul planar robotic arm is designed to create artistic, humanlike drawings on traditional media such as pen and paper [27]. Additionally, two-dimensional plotters such as Line-us and iDraw have become increasingly popular for automated writing tasks, demonstrating their ability to produce physically forged trajectories, as noted in [2]. Contrastingly, NAO robots serve a different functional purpose. These robots have been programmed effectively in various educational studies to assist children in enhancing their calligraphy skills, blending robotics with pedagogy [3]. Similarly, the UR5e robot, known for its precision and versatility, has been demonstrated to accurately replicate the trajectory and velocity of human signatures, making it a valuable tool in signature verification research [19]. We have also used UR5e in this work to further explore its application.

### 2.2. Machine learning to estimate function-based features

Recent advances in machine learning, particularly with neural networks, have significantly improved the analysis of function-based features across various domains, including bio- metrics and robotics. These models are especially effective in handling data characterized by complex temporal sequences and variability between executions. For instance, Recurrent Neural Networks have been widely applied to model time- dependent data across diverse fields, ranging from financial forecasting to motion analysis [13, 18, 28]. In biometrics, the predictive power of a hybrid CNN-GRU network has been used to effectively distinguish between Parkinsonian and healthy handwriting [8]. A similar approach has been used in robotics to estimate object trajectories for robot manipulation [22]. In this paper, however, we preferred a simple approach based on an MLP and a sliding window over features. Our goal was not specifically to advance state-of-the-art sequence modelling, but rather, to develop a model robust enough to estimate kinematics and dynamics effectively and apply them across different datasets.

### 2.3. Robotic features in signature verification

One advantage of our work is the estimation of robotic features based on the movement of the UR5e robotic arm when signing, which represents a novel approach in comparison to recent studies [1]. This method leverages real-world robotic data to enhance feature extraction, offering a novel perspective in the field. However, a limitation of our approach is that it relies on a single robot for all signature data. This means our method does not account for individual variations in arm and forearm anatomy among different writers, a factor not captured in the signature datasets.

Robotic features are increasingly being investigated in the context of signature verification to enhance the security capa- bilities of automatic verification systems. In [6], the authors explored the complexities of inter-joint motion in hand skeleton movements, which are challenging to reproduce. Their re- search focused on replay attacks involving a robotic hand repli- cating an in-air signature, highlighting the difficulties in accurately simulating human hand dynamics. Further studies have analysed the kinematics of a virtual skeleton arm during the signature execution process. An IRB120 ABB robotic model aimed at examining the angular position of the joints was used in [7], demonstrating significant improvements in signature verification performance through precise emulation of joint movements.

Investigations carried out thus far have primarily focused on describing the motion of joints, but information about the forces causing these movements has not been extensively explored. In the present work, we introduce the application of torque mea- surements, a novel approach that considers the force torques required to execute joint rotations in real robots during signa- ture execution. Previous studies [17, 20] introduced the concept of torques, which are derived from the geometric properties of written trajectories, to this field. In this article, torques refer specifically to the force torques required to execute rotations in the joints of real robots for signature execution.

Although features extracted from digitizers [23] or engineered features such as curvature and torsion values [25] are well-established, the features proposed in this work introduce a novel feature space in the field of signature verification.

## 3. Proposed method

Our method aims to estimate the kinematic and dynamic features of a signature, as depicted in Figure 1. First, we use the UR5e robotic arm to extract these features from signatures. Then, the features are used to train an MLP neural network. Once its weights are adjusted, the MLP is used to estimate these features from the trajectory of a new signature.

### 3.1. Kinematic and dynamic features with the robotic arm

We utilized the Universal Robots UR5e anthropomorphic robot arm[1] as the platform for replicating signatures and ac- quiring kinematic and dynamic features. The UR5e is commonly employed in industrial and service tasks requiring human collaboration. This robot arm can lift payloads of up to 5 kg, extend to a maximum reach of 850 mm, and operate with 6 degrees of freedom, enabling precise positioning of the end effector in any desired orientation and position.

The Forward Kinematics (FK) of the robotic kinetic chain was modelled using Denavit-Hartenberg (D-H) parameters [5]. These parameters define the relative positioning of consecutive elements within the chain. The relative poses between the frames of reference attached to the subsequent $(i-1)$-th and $i$-th links of the robot can be derived using the D-H parameters through four atomic movements:

$$^{i-1}_iT = Trot_z(\theta_i)Ttransl_z(d_i) \cdot Ttransl_x(a_i)Trot_x(\alpha_i),$$ (1)

where Trot and Ttransl represent rotation and translation trans formations around the z and x axes [4]. The full FK model of the UR5e arm can, therefore, be ex pressed as follows:

$$^b_eT = \prod^6_{i=1} {}^{i-1}_iT(a_i, \alpha_i, d_i, \theta_i),$$ (2)

where b denotes the base frame of the arm, while e is the end frame. The FK model of the robot calculates the pose of the end effector given the robot's configuration vector θr(t) = [θ1(t), . . . , θ6(t)]. Conversely, the Inverse Kinematics (IK) [14] involves determining the robot configuration θ when bT is given. We used the analytical model for the FK and the nu- merical solution for the IK.

Determining the joint torques $\tau$ needed to achieve the de- sired trajectory profile $[\theta(t), \dot{\theta}(t), \ddot{\theta}(t)]$ is the goal of Inverse Dynamics (ID). Accurate implementation often requires an ex- act knowledge of the robot's dynamic parameters, such as the mass distribution and inertia tensors, which are often inaccessible or inconsistently measured. Given that the UR5e robot lacks built-in force-torque sensors, we opted for estimating torques from motor currents during robot trajectory execution, such as $\tau_i = r_i \cdot K_{t,i} \cdot I_i$, where $\tau_i$ is the torque on the $i$-th motor, $r_i$ is the corresponding gear ratio, $K_{t,i}$ is the motor-specific torque-to-current coefficient, and $I_i$ is the current supplied to the motor. The $K_t$ coefficients, precalibrated by the robot manu- facturer, are retrievable from the robot controller's configuration files. For our experiments, the coefficients were set as $K_t$ = [0.1094, 0.1100, 0.1097, 0.0820, 0.0822, 0.0824], with gear ratios $r$ = [101, 101, 101, 101, 101, 101], as provided by the manufacturer in the calibration files.

The robot arm's control was achieved through a custom ROS package developed at Bialystok University of Technology[2]. We used a list of timed waypoints $[t, x, y, z]$ that were imbued with the desired pen orientation. The pen was assumed to align perpendicular to the writing surface. Next, we solved the IK task using a numerical Jacobian-based algorithm to establish the robot's list of timed configuration vectors $[t, \theta]$. We used the list of configurations to construct a joint-space spline inter- polator. This interpolator generates the immediate desired con- figurations, sent directly to the robot controller at a control fre- quency of 125 Hz. To achieve the desired accuracy, we had to modify the default robot controller parameters to $k_{gain} = 2000$ and $t_{lookahead} = 0.03$ s. The robot controller responded with the current joint state data, including the angular positions $\theta_r(t)$, ve- locities $\omega_r(t)$, and actual motor currents $I_r(t)$. We computed the torques $\tau_r(t)$ based on the currents.

### 3.2. Estimation of the kinematic and dynamic features with the neural network model

In order to find a more cost-effective alternative that eliminates the need for a physical robot, the objective is to develop a system capable of estimating the angular velocities θ(t), an gular positions ω(t), and force torques τ(t) of the robotic arm by leveraging the (x,y)-coordinates of points sampled from sig natures. This effort, framed as a multi-target regression, seeks to map the 2-dimensional space of signature coordinates, $\mathbb{R}^2$, to an 18-dimensional output space, $\mathbb{R}^{18}$,, where each dimension corresponds to one of the six values for θ(t), ω(t), and τ(t). The primary challenge of this task lies in accurately estimating these eighteen output values for each point based solely on the two input features despite the inherent variability in the lengths and styles of signatures.

The input features were normalized to the [0,1] range via min-max scaling, thereby maintaining generalizability across different datasets. Target values were subjected to a similar scaling process to the [0,1] range. Considering the broad spectrum of target values, this reversible scaling method preserves the model's general applicability and facilitates the learning process. Crucially, the parameters used for scaling were exclusively determined from the training datasets, ensuring that their subsequent application to the test sets would not result in information leakage.

We adopted an MLP for its adaptability in managing tasks that require multiple simultaneous outputs [29]. To enrich the model with sequential information, each signature point was not considered in isolation; instead, for each training in stance, we included the (x,y)-coordinates of the point itself, along with the coordinates of the five preceding and succeeding points. This sliding window approach, whose horizon was chosen based on its effectiveness in preliminary trials, incorporated contextual information surrounding each point, enhancing the MLP's ability to capture the dynamics of the signature move ment. Specifically, given an input vector x, the MLP performs a series of operations to estimate the outputs for $\hat{\theta}(t)$, $\hat{\omega}(t)$, and $\hat{\tau}(t)$. The architecture features a ReLU-activated hidden layer:

$$\mathbf{h} = \text{ReLU}\,(\boldsymbol{W}_h\mathbf{x}+\boldsymbol{b}_h) \tag{3}$$

with twelve units followed by a dropout layer (dropout rate of 0.3) to mitigate overfitting [24]:

$$\mathbf{h}' = \text{Dropout}(0.3, \mathbf{h}). \tag{4}$$

Uniquely, the model was structured with three separate output "heads", each comprising six units, facilitating the concurrent estimation of the three targets. Each output j (for j ∈ {θ,ω,τ}) is defined as:

$$\boldsymbol{y}_j = \sigma\,(\mathbf{W}_j\mathbf{h}' + \mathbf{b}_j), \tag{5}$$

where σ denotes the sigmoid activation function, ensuring that the estimated values are bounded within the [0,1] range. Our aim was to strike a balance between effectiveness and simplicity, grounded on common design choices. The designated loss function was a composite loss, calcu lated as the sum of three mean squared error (MSE) losses, each corresponding to one of the model's output heads. Mathematically, if $L_\theta$, $L_\omega$, and $L_\tau$ represent the MSE losses for the output heads dedicated to θ, ω, and τ respectively, the total loss $L_{total}$ is formulated as:

$$L_{total} = L_\theta + L_\omega + L_\tau \tag{6}$$

where $L_j = \frac{1}{n}\sum_{i=1}^{n}(\hat{y}_{i,j} - y_{i,j})^2$ for $j \in \{\theta, \omega, \tau\}$, $n$ is the number of samples, $\hat{y}_{i,j}$ is the estimated value, and $y_{i,j}$ is the true value for the $i^{th}$ sample of the $j^{th}$ output.

## 4. Experiments

The experiments aimed to analyse whether the features extracted from the MLP are similar to those extracted from the UR5e, both in terms of estimation and performance in ASV.

### 4.1. Datasets

Six signature databases were used to extract dynamic and kinematic features, train the MLP network, and validate the generalization of our proposed model. First, we used the MCYT330 sub-corpus [21] (DS1), which consists of signatures from 330 users, each providing 25 genuine and 25 forged signatures across two sessions. The BiosecurID [12] (DS2) multimodal database contains online signature data from 132 users. Each user contains 16 genuine and 12 skilled forgeries. The Visual [15] (DS3) sub-corpus includes data from 94 users, each providing 20 genuine signatures from two sessions and 10 skilled forgeries. The Blind sub-corpus [15] (DS4) includes 88 users, each providing 10 genuine signatures and 10 skilled forgeries. Finally, the OnOffSigDevanagari-75 (DS5) and OnOffSigBengali-75 (DS6) datasets [10] each have 75 users, with 24 genuine signatures per user. These databases provided diverse environments to test the adaptability and effectiveness of our method across different signature capture technologies and forgery scenarios.

### 4.2. Automatic signature verification

The automatic signature verifier developed in this study comprises two key components: function-based features and the verification mechanism itself.

The function-based features were organized into matrices with $N$ columns and $M$ rows. These columns represent concatenated function-based features that describe various aspects of the signature, while the rows correspond to the number of sampling points, which vary depending on the digitizer's frequency and the signature's length and duration. As such, while the number of columns was fixed across all experiments, the number of rows fluctuated with each signature. Following [7], each feature within the matrix was first normalized within the range [0, 1]. To enrich the feature set, each feature's first and second derivatives were calculated using a second-order regression and incorporated into the feature matrix, which was then standardized using z-score normalization.

The UR5e robot provided six initial function-based features for each degree of freedom. In the case of the angular posi- tion $\theta_r(t)$, a comprehensive feature matrix was developed. The same approach was applied to angular velocity $\omega_r(t)$ and force torques $\tau_r(t)$, and to their estimation provided by the MLP model. Notably, since the pen attached to the axis of the sixth joint of the UR5e did not rotate around its axis, the angular velocity $\omega_6(t)$ remained constant, and was consequently excluded from the verification process.

For classification, a standard implementation of the Dynamic Time Warping (DTW) [11] algorithm was used to optimize the Euclidean distance between two feature matrices. The computation was accelerated by applying the Sakoe-Chiba band [11], limiting the search width to $M/10$ of the diagonal. The verification process involves comparing a questioned signature $q$ against a set of $R$ reference signatures. The minimum DTW distance quantifies the relationship of $q$ to the reference set:

$s_R(q) = \arg\min_{r \in R}[DTW(q, r)]$, where $r$ is a reference signature, and $s_R(q)$ is the non-normalized score.

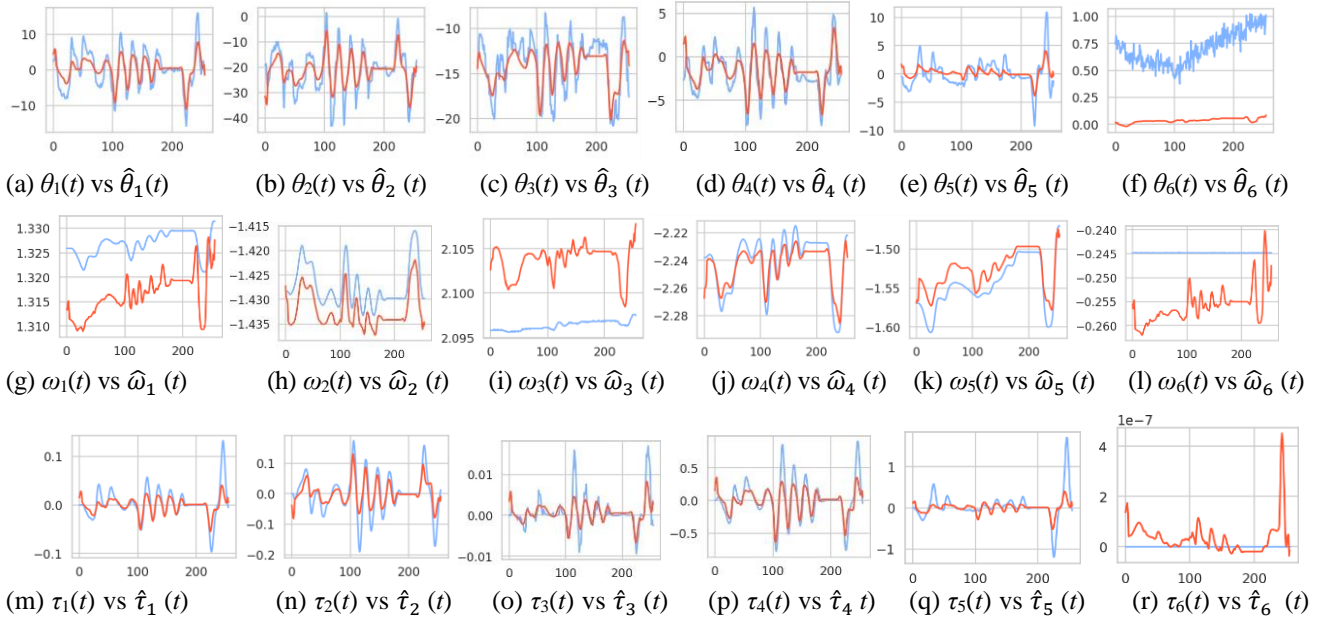

(a) $\theta_1(t)$ vs $\hat{\theta}_1(t)$    (b) $\theta_2(t)$ vs $\hat{\theta}_2(t)$    (c) $\theta_3(t)$ vs $\hat{\theta}_3(t)$    (d) $\theta_4(t)$ vs $\hat{\theta}_4(t)$    (e) $\theta_5(t)$ vs $\hat{\theta}_5(t)$    (f) $\theta_6(t)$ vs $\hat{\theta}_6(t)$

(g) $\omega_1(t)$ vs $\hat{\omega}_1(t)$    (h) $\omega_2(t)$ vs $\hat{\omega}_2(t)$    (i) $\omega_3(t)$ vs $\hat{\omega}_3(t)$    (j) $\omega_4(t)$ vs $\hat{\omega}_4(t)$    (k) $\omega_5(t)$ vs $\hat{\omega}_5(t)$    (l) $\omega_6(t)$ vs $\hat{\omega}_6(t)$

(m) $\tau_1(t)$ vs $\hat{\tau}_1(t)$    (n) $\tau_2(t)$ vs $\hat{\tau}_2(t)$    (o) $\tau_3(t)$ vs $\hat{\tau}_3(t)$    (p) $\tau_4(t)$ vs $\hat{\tau}_4(t)$    (q) $\tau_5(t)$ vs $\hat{\tau}_5(t)$    (r) $\tau_6(t)$ vs $\hat{\tau}_6(t)$

Fig. 2: Visual representation of the sequences of angular positions $\theta(t)$, angular velocities $\omega(t)$, and force torques $\tau(t)$ for signature 005f01 of DS1. The sequences obtained by the UR5e are shown with a blue line, while those estimated by the MLP are shown with a red line.

Two stages of normalization were subsequently applied to these scores. The first stage uses the warping path length $|p|$ to normalize scores for detecting random forgeries: $\hat{s}_{R1}(q) = s_R(q)/|p|$. In the second stage, a weighted factor $\mu_R$, derived from all reference signatures of the user, is used to normalize scores for detecting skilled forgeries: $\hat{s}_{R2}(q) = s_R(q)/\mu_R$.

### 4.3. Experimental setup and metrics

For feature estimation using the MLP, we employed a four-model approach to estimate each signature in the DS1 dataset. Each model used identical hyperparameters, but was tested on distinct quarters of the dataset. The outputs from these models were averaged to produce the final results. To assess generalizability, a final model was trained on the entire DS1 dataset. Training utilized the Adam optimizer [16] with a commonly used learning rate of 0.01. An early stopping mechanism based on the validation loss observed on 20% of the training set was employed to prevent overfitting. The patience parameter was set to 1 epoch, leading to convergence within a few epochs. Performance metrics included the mean absolute error (MAE) and mean squared error (MSE).

For signature verification, we approached the *random forgeries* and *skilled forgeries* experiments, following the common standard of the ICDAR 2021 signature competition [26]. We used five random genuine signatures from each writer as reference signatures, while the remaining genuine signatures served as test samples. For random forgery, one genuine signature from each user was randomly selected from all other users. As recommended in [26], all the skilled forgeries were used in the test, but not employed as reference negative signatures. Finally, to illustrate the system's efficacy at various False Acceptance Rate (FAR) and False Rejection Rate (FRR) levels, the Detection Error Trade-off (DET) curves were shown on a logarithmic scale. As is common in signature verification [9], we used the Equal Error Rate (EER) to evaluate the performance. Note that the EER represents the point on the DET plot where the FAR and FRR are equal. These experiments were conducted ten times, the DET curves were averaged, and the EER is reported as the average ±standard deviation for each case.

Table 1: UR5e kinematic and dynamic estimation performances.

| Model | Parameter | MAE | MSE |
|-------|-----------|-----|-----|
| MLP | Angular velocities, $\theta(t)$ | 0.0041 | 0.0001 |
| | Angular positions, $\omega(t)$ | 0.0571 | 0.0063 |
| | Force torques, $\tau(t)$ | 0.0126 | 0.0004 |
| RNN | Angular velocities, $\theta(t)$ | 0.1237 | 0.0663 |
| | Angular positions, $\omega(t)$ | 0.0276 | 0.0015 |
| | Force torques, $\tau(t)$ | 3.1420 | 23.2520 |
| LSTM | Angular velocities, $\theta(t)$ | 0.1283 | 0.0694 |
| | Angular positions, $\omega(t)$ | 0.0268 | 0.0015 |
| | Force torques, $\tau(t)$ | 3.0116 | 20.1721 |
| GRU | Angular velocities, $\theta(t)$ | 0.1370 | 0.0800 |
| | Angular positions, $\omega(t)$ | 0.0285 | 0.0016 |
| | Force torques, $\tau(t)$ | 3.8371 | 34.4615 |

### 4.4. Feature estimation results

The UR5e kinematic and dynamic estimation results are summarized in Table 1. To explore alternatives, we show the results obtained not only with the proposed MLP model but also with state-of-the-art techniques, including Recurrent Neural Network (RNN), Long Short-Term Memory (LSTM), and Gated Recurrent Unit (GRU) models, which are well-known for their effectiveness in sequence prediction tasks.

Specifically, the RNN, LSTM, and GRU architectures employed in this study consist of a bidirectional layer with 32 units followed by a time-distributed layer with 18 output units, corresponding to the six values for $\theta(t)$, $\omega(t)$, and $\tau(t)$. The experimental setup for these models was the same as for the MLP to ensure a fair comparison. The only difference was the additional preprocessing step needed to pad the sequences to ensure uniform length, as these models accept sequences as input in- stead of sliding windows of points.

The findings reveal distinct differences in model performance across the various parameters. The MLP significantly outperforms the RNN, LSTM, and GRU models in estimating angular velocities and force torques, with much lower MAE and MSE values. This suggests that the sliding window approach used in the MLP effectively captures the local context of signature points, which is crucial for accurately estimating these parameters. Conversely, the sequential models may struggle to capture this fine-grained context, leading to higher errors. For angular positions, the RNN, LSTM, and GRU models exhibit slightly better performance than the MLP. This indicates that angular positions benefit from the sequential nature of these models, which can better capture temporal dependencies in the data.

Overall, the results showcase high accuracy in the model's estimations, especially with regard to angular velocities and force torques. However, numerical results alone do not fully allow to evaluate the model's proficiency in accurately capturing the underlying trends over time. To address this, Figure 2 visually compares the model's estimated trends against actual data. The model demonstrates high accuracy in estimating an- gular velocities and force torques. However, there are some mi- nor deviations in amplitude and fluctuations along the *y*-axis. The sixth value for the angular velocities and force torques is challenging to estimate as the actual values remain relatively constant. Although the model shows an overall accurate replication of the function behaviour for angular positions, there is a slight offset. In summary, the model is somewhat accurate, but does not always perfectly mimic the function behaviour.

*4.5. Online signature verification results*

The effectiveness of the method is demonstrated in two ways. First, Figure 3 shows that features generated by the neural network perform similarly to, or even better than, those obtained from the robot at the EER level. In the figure, gray lines represent baseline results using function-based features directly extracted from the UR5e robot: angular positions, $\theta$, are depicted with solid lines, angular velocities, $\omega$, with dashed lines, and torques, $\tau$, with dotted lines. The blue lines represent performance using estimated features with the same database (DS1). Comparing the results for each type of features, we observe al- most similar results on $\theta$. However, in the case of $\omega$ and $\tau$, the EER was always better when these sequences were extracted from the neural network. Also, corresponding DET curves from MLP are always below the UR5e ones. These results satisfy our requirements with the neural network, which are consistent in both random and skilled forgeries.

It could be said that al- though the UR5e robotic arm provides precise measurements, it can result in overfitting while performing signature verification. Moreover, the MLP model is designed to generalize better from training data to unseen data, even though it uses estimated features. This improves its performance and helps prevent over- fitting.

Second, Figure 4 examines the generalization capacity of the MLP model, trained solely with data from DS1, and further tested with signatures from DS2 to DS6, which were never processed by the UR5e robot. This created a challenging situation for evaluating the utility of the estimated features. The findings revealed that in DS2, while there was a significant decline in EER for $\theta$ in random forgeries, $\omega$ and $\tau$ delivered results comparable to DS1. Similar outcomes were observed in skilled forgeries for DS2, with angular velocities and force torques showing improved EERs. For DS3, the angular position again displayed outlying results in random forgeries; however, angular velocities and torques performed slightly better. In skilled forgeries, stable EERs comparable to DS2 were achieved with angular velocities and torques. For DS4, we observed better performance for $\theta$ in both random and skilled forgeries, with slightly higher results for skilled forgeries in angular velocities and torques. DS5 and DS6 reported similar performance in random forgeries, consistent with the other results. Note that these two datasets did not include skilled forgeries. These results underscore the robustness of the angular velocity and force torques obtained by the MLP across different datasets and forgery scenarios.

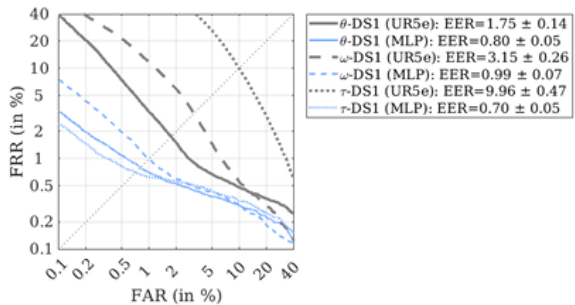

(a) Random forgeries

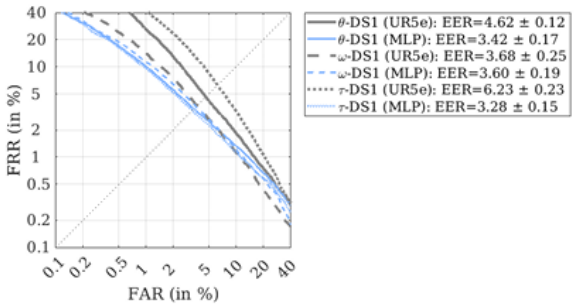

(b) Skilled forgeries

Fig. 3: Comparison of performance between the baseline using UR5e features and estimated features, illustrated with DET plots and EER for signature verification results.

## 5. Conclusions and future works

We introduced a neural network model to estimate the kinematic and dynamic features of a robotic arm. We trained the model using real robotic features and demonstrated its practical application in signature verification. The MLP demonstrated the ability to estimate both kinematic and dynamic features from the trajectory of unseen signatures. Furthermore, the results also showed that the neural network used can reduce the need for experimentation with a real robot to extract features due to its generalization in ASV. The robotic features and the model are publicly available for further research.

For future works, one promising direction is to develop a custom robotic arm tailored specifically for signature verification, potentially replicating the kinematics and dimensions of the human arm based on anthropometric data. It could enhance the accuracy of signature analysis by capturing the nuances of signature dynamics more precisely. A potential limitation of our work is the assumption that the same robot configuration is suitable for all signers. While personalizing the robot model for each signer might better capture individual human variability, it is uncertain whether this would actually improve the performance of the automatic signature recognizer. Additionally, investigating interpretable neural network models could offer valuable insights into the decision-making process. Further improvements include attention mechanisms to refine model performance. These advancements could significantly improve the effectiveness and adaptability of the signature verification system.

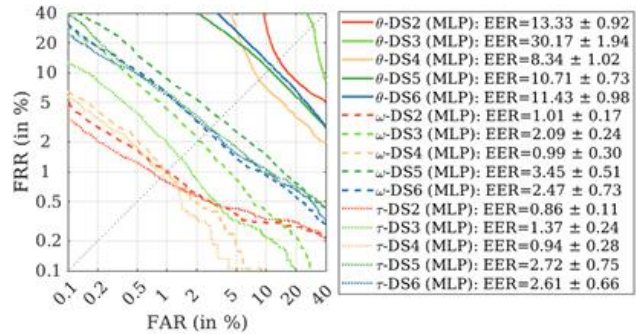

(a) Random forgeries

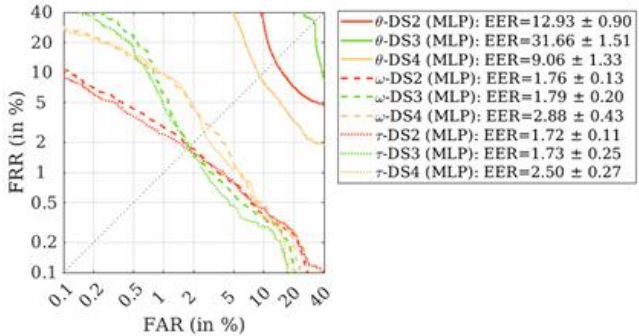

(b) Skilled forgeries

Fig. 4: Performance results across different databases, trained with DS1 data, illustrated through DET plots and EER values for signature verification.